\pdfoutput=1 
\documentclass[letterpaper, 10 pt, conference]{ieeeconf}  

\IEEEoverridecommandlockouts                              

\useRomanappendicesfalse

\usepackage{amsmath}
\usepackage{bm}
\usepackage{amsfonts}
\usepackage{mathtools}
\usepackage{color}

\usepackage{tabularx,booktabs}
\usepackage{threeparttable}
\usepackage{multirow}

\newcommand{\xx}{\bm{x}}

\newcommand{\varparams}{\phi}

\newcommand{\varbound}{\mathcal{L}}

\newcommand{\yy}{\bm{y}}
\newcommand{\zz}{\bm{z}}

\newcommand{\probvec}{\bm{p}_{\yy | \zz}}

\newcommand{\defeq}{\vcentcolon =}
\newcommand{\sigmoid}{\mathsf{Sigmoid}}
\newcommand{\bernoulli}{\mathsf{Bern}}

\newcommand{\hip}[1]{\textcolor{blue}{\textbf{Ingmar}: #1}}
\newcommand{\graveyard}[1]{}

\usepackage{graphicx}
\usepackage{float}
\graphicspath{ {./figures/} }
\usepackage{lipsum}

\usepackage[caption=false]{subfig}

\usepackage[normalem]{ulem}
\useunder{\uline}{\ul}{}

\title{\LARGE \bf
Probably Unknown: Deep Inverse Sensor Modelling In Radar}

\author{Rob Weston, Sarah Cen, Paul Newman and Ingmar Posner
\thanks{Authors are from the Oxford Robotics Institute (ORI)}
\thanks{\texttt{\{robw,sarah,pnewman,ingmar\}@robots.ox.ac.uk}}}

\begin{document}

\setlength{\abovedisplayskip}{3pt}
\setlength{\belowdisplayskip}{2.9pt}

\renewcommand{\baselinestretch}{0.85}

\maketitle
\thispagestyle{empty}
\pagestyle{empty}

\begin{abstract}
Radar presents a promising alternative to lidar and vision in autonomous vehicle applications, able to detect objects at long range under a variety of weather conditions. However, distinguishing between occupied and free space from raw radar power returns is challenging due to complex interactions between sensor noise and occlusion.

To counter this we propose to learn an Inverse Sensor Model (ISM) converting a raw radar scan to a grid map of occupancy probabilities using a deep neural network. Our network is self-supervised using partial occupancy labels generated by lidar, allowing a robot to learn about world occupancy from past experience without human supervision. We evaluate our approach on five hours of data recorded in a dynamic urban environment. By accounting for the scene context of each grid cell our model is able to successfully segment the world into occupied and free space, outperforming standard CFAR filtering approaches. Additionally by incorporating heteroscedastic uncertainty into our model formulation, we are able to quantify the variance in the uncertainty throughout the sensor observation. Through this mechanism we are able to successfully identify regions of space that are likely to be occluded.

\end{abstract}

    

\section{INTRODUCTION}
Occupancy grid mapping has been extensively studied \cite{elfes1989using,konolige1997improved} and successfully utilised for a range of tasks including localisation \cite{milstein2008occupancy,filliat2003map} and path-planning \cite{meyer2003map}. 
One common approach to occupancy grid mapping uses an inverse sensor model (ISM) to predict the probability that each grid cell in the map is either \textit{occupied} or \textit{free} from sensor observations. Whilst lidar systems provide precise, fine-grained measurements, making them an obvious choice for grid mapping, they fail if the environment contains fog, rain, or dust \cite{clarke2013towards}. Under these and other challenging conditions, FMCW radar is a promising alternative that is robust to changes in lighting and weather and detects long-range objects, making it well suited for use in autonomous transport applications. 

However, two major challenges must be overcome in order to utilise radar to this end. Firstly, radar scans are notoriously difficult to interpret due to the presence of several pertinent noise artefacts. Secondly, by compressing information over a range of heights onto a dense 2D grid of power returns identifying occlusion becomes difficult. The complex interaction between occlusion and noise artefacts introduces uncertainty in the state of occupancy of each grid cell which is \emph{heteroscedastic}, varying from one world location to another based on scene context, and \emph{aleatoric} \cite{kendall2017uncertainties}, inherent in radar data by way of the scan formation process. 

In order to successfully reason about world occupancy, we posit that a model that is able to reason about scene context is essential. To this end, we formulate the problem of determining an ISM as a segmentation task, leveraging a deep network to learn the probability distribution of occupancy from raw data alone. This allows us to successfully determine regions of space that are likely to be occupied and free in light of challenging noise artefacts. Simultaneously, by explicitly modelling heteroscedastic uncertainty, we are able to quantify the latent uncertainty associated with each world cell arising through occlusion. Utilising approximate variational inference we are able to train our network using self-supervision relying on partial labels automatically generated from occupancy observations in lidar.

We train our model on real-world data generated from five hours of urban driving and successfully distinguish between occupied and free space, outperforming constant false-alarm rate (CFAR) filtering in average intersection over union performance. Additionally we show that by modelling heteroscedastic uncertainty we are able to successfully quantify the uncertainty arising through the occlusion of each grid cell.

\begin{figure}
    \centering
    \vspace{0cm}
    \includegraphics[width=\linewidth]{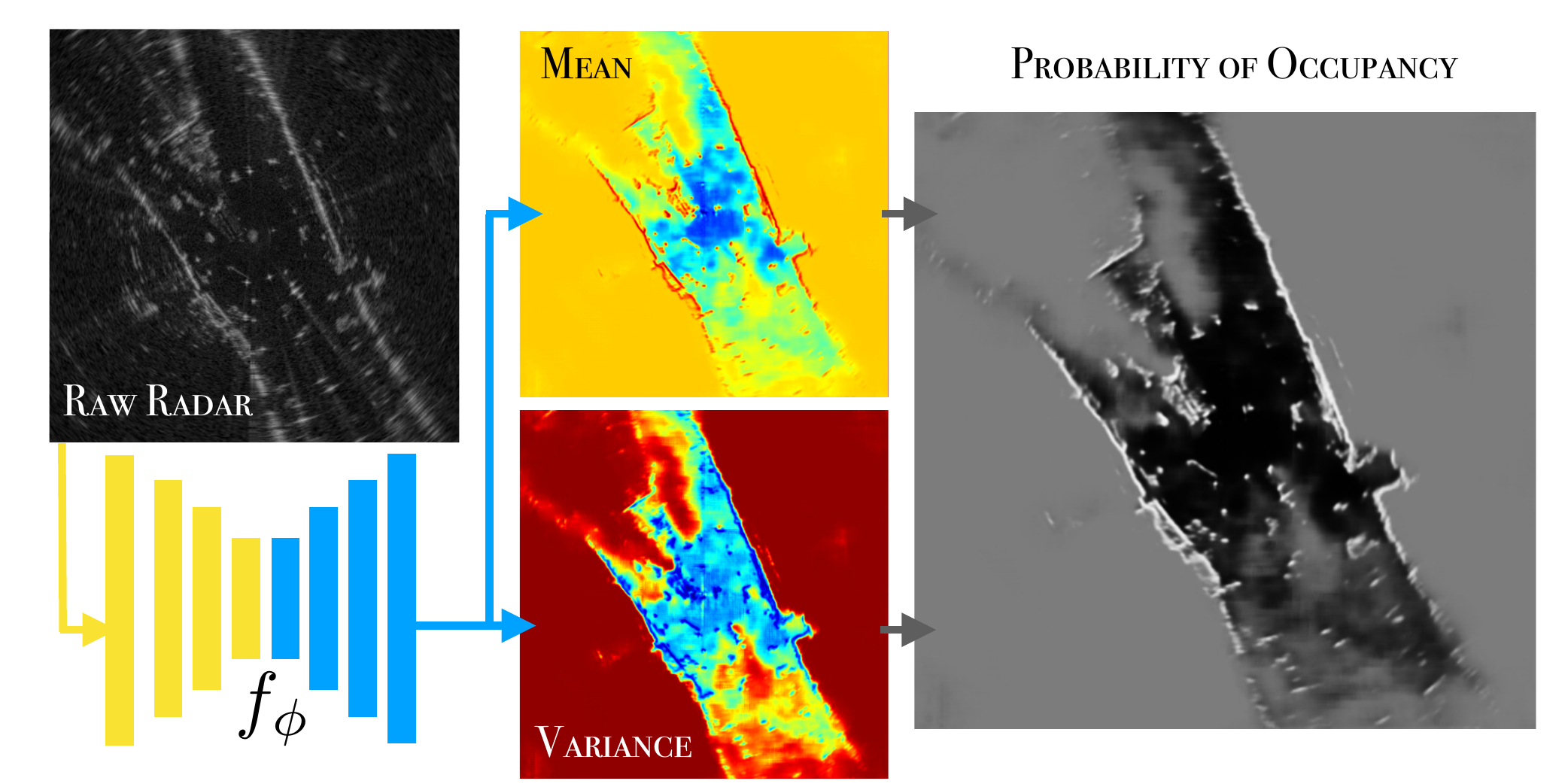}
    \vspace{-0.5cm}
    \caption{Our network learns the distribution of occupancy from experience alone. By reasoning about scene context it is able to successfully identify regions of space that are likely to be occupied and free. The uncertainty associated with each grid cell is allowed to vary throughout the scene by predicting the noise standard deviation alongside the predicted logit of each grid cell. These are combined to generate a grid map of occupancy probabilities. The uncertainty predicted by our network can be used to successfully identify regions of space that are likely to be occluded.}
    \label{fig:taesar}
    \vspace{-0.5cm}
\end{figure}

\section{RELATED WORK}
Inverse sensor models (ISMs) \cite{elfes1989using} are used to convert noisy sensor observation to a grid map of occupancy probabilities. For moving platforms, a world occupancy map can then be sequentially generated from an ISM, multiple observations, and known robot poses using a binary Bayes filter \cite{thrun2005probabilistic}. Using lidar data, ISMs are typically constructed using a combination of sensor-specific characteristics, experimental data, and empirically-determined parameters \cite{werber2015automotive,dia2017evaluation,thrun2003learning}. These human-constructed ISMs struggle to model challenging radar defects and often utilise limited local information to predict each cell's occupancy without accounting for scene context. 

Instead, raw radar scans are often naively converted to binary occupancy grids using classical filtering techniques that distinguish between objects (or targets) and free space (or background). Common methods 
include CFAR \cite{skolnik2008radar} and static thresholding. However, both return binary labels rather than probabilities, and neither is capable of addressing all types of radar defects or capturing occlusion. Additionally, the most popular approach, CFAR, imposes strict assumptions on the noise distribution and requires manual parameter tuning. In contrast, using deep learning methods, as first proposed by \cite{thrun1993exploration}, allows the distribution of world occupancy to be learned from raw data alone, accounting for the complex interaction between sensor noise and occlusion through the higher level spatial context of each grid cell.

In order to capture uncertainty that varies from one grid cell to the next we incorporate heteroscedastic uncertainty into our formulation inspired by \cite{kendall2017uncertainties}. Our variational re-formulation of \cite{kendall2017uncertainties} is closely related to the seminal works on variational inference in deep latent variable models \cite{rezende2014stochastic, kingma2013auto} and their extension to conditional distributions \cite{sohn2015learning}.

Drawing on the successes of deep segmentation in  biomedical applications, \cite{roth2018deep} and vision \cite{liu2018recent} we reformulate the problem of learning an inverse sensor model as neural network segmentation. Specifically, we utilise a U-net architecture with skip connections \cite{ronneberger2015u}. In order to map from an inherently polar sensor observation to a Cartesian map we utilise Polar Transformer Units (PTUs) \cite{esteves2017polar}.

\section{DEEP INVERSE SENSOR MODELLING IN RADAR}
\subsection{Setting}
\label{sec:setting}
\begin{figure}[h]
    \vspace{0.15cm}
    \centering
    \includegraphics[width=1\linewidth]{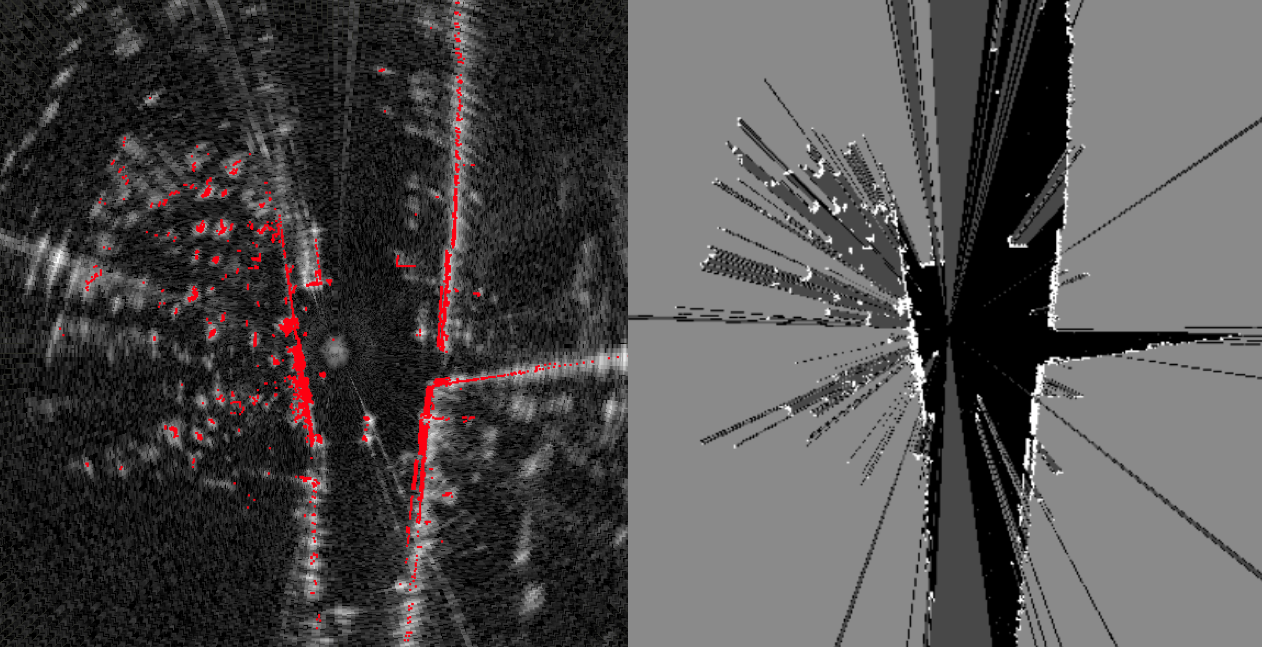}
    \vspace{-1.3em}
    \caption{Generated training labels from lidar. The image on the left shows the lidar points (red) projected into a radar scan $\xx$ converted to Cartesain co-ordinates for visualisation. The right image shows the generated training labels. Any grid cell $(u,v)$ with a lidar return is labelled as occupied $\hat{\yy}^{u, v}=1$ (white). Ray tracing along each azimuth, the space immediately in front of the first return is labelled as $\hat{\yy}^{u, v} = 0$ (black), the space between the first and last return or along azimuths in which there is no return is labelled as \textit{partially observed}, $\bm{o}^{u,v}=2$, (dark grey) and the space behind the last return is labelled as \textit{unobserved}, $\bm{o}^{u,v}=0$, (light grey). Any space that is labelled as occupied or free is labelled as $\textit{observed}$, $\bm{o}^{u,v}=1$}
    \label{fig:generating_training_labels}
    \vspace{-0.4cm}
\end{figure}

Let $\xx \in \mathbb{R}^{\Theta \times R}$ denote a full radar scan containing  $\Theta$ azimuths of power returns at $R$ different ranges for each full rotation of the sensor. Partitioning the world into a $H \times W$ grid, $\yy \in \{0, 1\}^{H \times W }$ gives the occupancy state of each grid cell, where $\yy^{u,v} = 1$ if cell $(u, v)$ is \textit{occupied} and $\yy^{u,v} = 0$ if $(u, v)$ is \textit{free}. \textit{Partial} measurements of occupancy $\hat{\yy}$ are determined by combining the output of multiple 3D lidars and projecting the returns over a range of heights onto a 2D grid. In order to separate the region of space where no labels exist most likely as a consequence of full occlusion, from space that is likely to only be partially occluded or for which no labels exist due to a limited field of view of the lidar sensors, the observability state of each cell $\bm{o}^{u, v}$ is recorded as 0, 1 or 2 corresponding to \textit{unobserved}, \textit{observed} and \textit{partially observed} space respectively. The full labelling procedure is described in Figure \ref{fig:generating_training_labels}. This process is repeated for $N$ radar-laser pairs to generate a data set $\mathcal{D} = \{\xx^{n}, (\hat{\yy}, \bm{o})^{n} \}_{n=1}^N$ of training examples from which we aim to learn an inverse sensor model $\bm{p}_{\yy | \xx} \in [0, 1]^{H \times W }$ such that $\bm{p}_{\yy | \xx}^{u, v} = p(\yy^{u, v} = 1 | \xx) $ gives the probability that cell $(u, v)$ is occupied dependent on the \textit{full} radar scan $\xx$

\subsection{Heteroscedastic Aleatoric Uncertainty and FMCW Radar}
\begin{figure*}
    \centering
    \includegraphics[width=0.95\linewidth]{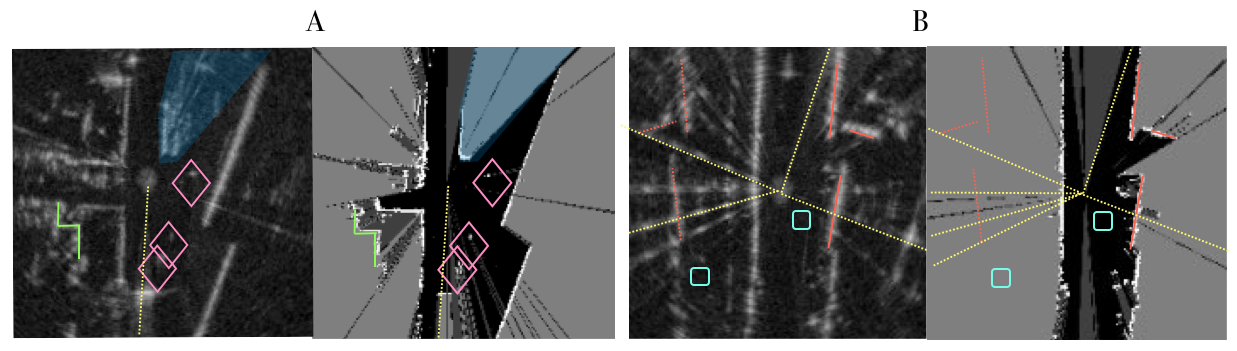}
    \vspace{-1em}
    \caption{Raw radar and the lidar ground truth. An ISM must be able to pick out faint objects, such as cars (pink diamonds), from the background speckle noise, in light of challenging noise artefacts such as saturation (yellow lines). In addition, an ISM must be able to determine which regions of space are likely to be occluded such as the space behind buses (highlighted blue) in light of almost identical local appearances (blue cyan boxes). Finally an ISM should be able to distinguish ghost objects (dotted orange) from true second returns (green lines).}
    \label{fig:radar_challenges}
    \vspace{-0.7cm}
\end{figure*}

FMCW Radar is an inherently noisy modality suffering from speckle noise, phase noise, amplifier saturation and ghost objects. These conspire to make the distinction between occupied and free space notoriously difficult. A radar's long range as well as its ability to penetrate past first returns make it attractive but also challenging. In particular, a radar's capacity for multiple returns along an azimuth implies varying degrees of uncertainty depending on scene context: the distinction between occupied and free space becomes increasingly uncertain as regions of space become partially occluded by objects. Examples of each of these problems are further explained in Figure \ref{fig:radar_challenges}. As such, high power returns do not always denote occupied and likewise, low power returns do not always denote free.

\graveyard{
\hip{is the remainder of this para needed here or can we leave it for figure caption?} For example, the cyan squares in Figure \ref{} show two regions of space one of which is fully occluded by the wall in front of it, whilst the other shows a region of space that is free. In the raw radar image both regions look almost identical. Similarly, the region highlighted blue in Figure \ref{fig:radar_challenges}, corresponds to space occluded as a bus passes close to the sensor, fully occluding the scene behind it. In contrast, the green lines in Figure \ref{fig:radar_challenges}b show second (and third)  returns corresponding to actual walls in the scene, as can be seen in the laser. }

Uncertainties in our problem formulation depend on the world scene through a complex interaction between scene context and sensor noise, and are inherent in our data as a consequence of the image formation process. As such they are, heteroscedastic as they depend on scene context and aleatoric as they are ever present in our data \cite{kendall2017uncertainties}. In order to successfully determine world occupancy from an inherently uncertain radar scan we seek a model that explicitly captures heteroscedastic aleatoric uncertainty. By framing this problem as a deep segmentation task we leverage the power of neural networks to learn an ISM which accounts for scene context in order to determine -- from raw data alone -- \emph{occupied} from \emph{free} space in the presence of challenging noise artefacts. Simultaneously, as a result of our heteroscedastic uncertainty formulation we are also able to learn which regions of space are inherently uncertain because of occlusion.

\subsection{Modelling Heteroscedastic Aleatoric Uncertainty}
Instead of assuming that the uncertainty associated with each grid cell is fixed, as is typically assumed in standard deep segmentation approaches, by using a heteroscedastic model the uncertainty in each grid cell $\bm{\gamma}_\varparams(\xx)$ is allowed to vary. This is achieved by introducing a normally distributed latent variable $\zz^{u, v}$ associated with each grid cell \cite{kendall2017uncertainties} and predicting the noise standard deviation $\bm{\gamma}_\varparams(\xx)$ alongside the predicted logit $\bm{\mu}_\varparams(\xx)$ of each each $\zz^{u, v}$ with a neural network $f_\varparams$ :

\begin{gather}
    p_\varparams(\zz | \xx) = \mathcal{N}(\zz | \bm{\mu}_\varparams(\xx), \bm{\gamma}_\varparams(\xx) \bm{I}) \\
    [\bm{\mu}_\varparams(\xx),  \bm{\gamma}_\varparams(\xx)] \defeq f_\varparams(\xx)
\end{gather}

Assuming a likelihood $p(\yy^{u, v} = 1 | \zz^{u, v}) = \sigmoid(\zz)$, the probability that cell $\yy^{u, v}$ is occupied is then given by marginalising out the uncertainty associated with $\zz$:
\begin{gather}
    \label{eq:cell_posterior}
    p(\yy^{u, v}| \xx) = \int p(\yy^{u, v} | \zz^{u, v}) p_{\varparams}(\zz^{u, v} | \xx)d \zz^{u, v}
\end{gather}

Unfortunately the integral in (\ref{eq:cell_posterior}) is intractable and is typically approximated using Monte-Carlo sampling and the reparameterization trick \cite{kendall2017uncertainties}. Instead, by introducing an analytic approximation in Section \ref{sec:inferenc} we show that we can accurately and efficiently approximate (\ref{eq:cell_posterior}) without resorting to sampling.

One final problem remains. We expect our model to be inherently uncertain in occluded space for which no lidar training labels are available. How do we train $f_\phi$ whilst explicitly encoding an assumption that in the absence of training labels we expect our model to be uncertain? In Section \ref{sec:training_with_partial} we propose to solve this problem by introducing a normally distributed prior $p(\zz)$ on the region of space for which no training labels exist utilising the variational inference framework.

\subsection{Training with Partial Observations}
\label{sec:training_with_partial}
In order to encode an assumption that in the absence of training data we expect our model to be explicitly uncertain we introduce a prior $p(\zz) = \mathcal{N}(\zz | \bm{\mu}, \bm{\gamma} \bm{I})$ on the uncertainty associated with the occluded scene which our network reverts back to in the absence of a supervised training signal. To do this, we begin by treating $p_{\varparams}(\zz | \xx)$ as an approximate posterior to $p(\zz | \yy)$ induced by the joint $p(\zz, \yy) = p(\yy | \zz) p(\zz)$ where,
\begin{gather}
\quad p(\yy | \zz) \defeq \prod_{u, v} \bernoulli (\yy^{u, v}| \bm{p}_{\yy | \zz}^{u,v}) \\
\bm{p}_{\yy | \zz}^{u,v} = p(\yy^{u, v} = 1) = \sigmoid (\zz^{u, v}) \\
    \quad p(\zz) \defeq \mathcal{N}(\zz | \bm{\bm{0}}, \gamma I)
\end{gather}
\noindent $\sigmoid$ and $\bernoulli(y | p) = p^{y}(1-p)^{1-y}$ denote the element-wise sigmoid function and Bernoulli distribution.

Next given a set of observations $\mathcal{D}$, we consider determining our parameters $\phi$ by maximising the variational lower bound,
\begin{gather}
\label{eq:true_varbound}
\varbound(\varparams; \mathcal{D}) =  \sum_n \mathcal{L}^n(\phi)  \\
         \mathcal{L}^n(\varparams) =  \mathbb{E}_{p_\varparams(\zz | \xx^n)}[\log p(\yy^n | \zz)] - d_{kl}[p_\varparams(\zz | \xx^n) || p(\zz)] \label{eq:varbound} 
\end{gather}
\noindent where $d_{kl}$ denotes KL divergence. The first term in $\mathcal{L}^n(\varparams)$ is the expected log-likelihood under the approximate posterior  ${p_\varparams(\zz | \xx)}$ which, when optimised, forces the network to maximise the probability of each occupancy label $\yy$. The second term forces ${p_\varparams(\zz | \xx)}$ towards the prior $p(\zz)$. 

Crucially, by only evaluating the log-likelihood term in the labelled region of space and only evaluating the KL divergence term in occluded space, we are able to train our network to maximise the probability of our labels whilst explicitly encoding an assumption that in the absence of training labels we expect our network to be inherently uncertain. The latter is achieved by setting the prior to $p(\zz) = \mathcal{N}(\zz | \bm{\bm{0}}, \gamma I)$ corresponding to an assumption that occluded space is equally likely to be free or occupied with a fixed uncertainty $\gamma$. We tested multiple values of $\gamma$ and found that setting $\gamma=1$ gave good results.

For a Gaussian prior and approximate posterior the KL divergence term can be determined analytically, whilst the expected log-likelihood is estimated using the reparameterization trick \cite{kingma2013auto} by sampling $\zz^{l} = \bm{\mu}_\varparams(\xx) + \bm{\gamma}_\varparams(\xx) \circ \bm{\epsilon}^l$ where $\bm{\epsilon}^l \sim \mathcal{N}(\bm{0}, \bm{I})$. The expected log-likelihood is then approximated as $\mathbb{E}_{p_\varparams(\zz | \xx)}[\log p(\yy | \zz)] \approx  -\frac{1}{L} \sum_{l} \big(\sum_{u,v} \mathbb{H}[\yy^{u, v}, \probvec^{l, u, v}] \big)$  where $\mathbb{H}$ denotes binary cross entropy.

Finally our loss function becomes
\begin{multline}
\label{eq:final_loss}
    \hat{\mathcal{L}}^n(\phi) = \frac{\bar{\omega}}{L} \sum_{l, u, v} \mathbb{I}(\bm{o}^{u,v}=1) \mathbb{H}_{\alpha}[\hat{\yy}^{n, u, v}, \probvec^{n, l, u, v}] \\ + \sum_{u, v} \mathbb{I}(\bm{o}^{u,v}=0) d_{kl}[p_\varparams(\zz^{u,v} | \xx^n)|| p(\zz^{u, v})] 
\end{multline}
\begin{equation}
    \hat{\mathcal{L}}(\phi; \mathcal{D}) = \frac{1}{N} \sum_n \hat{\mathcal{L}}^n(\phi)
\end{equation}
\noindent where $\mathbb{I}$ denotes the indicator function which is equal to $1$ if its condition is met and $0$ otherwise.

In order to ensure that labelled and unlabelled data contribute equally to our loss we re-weight the likelihood term with $\bar{\omega} = \omega HW / (\sum_{uv} \mathbb{I}(\bm{o}^{u,v}=1))$. The hyper-parameter $\omega$ is used to weight the relative importance between our prior and approximate evidence. As there is also a significant class imbalance between occupied and free space we use weighted binary cross entropy $\mathbb{H}_{\alpha}$ where the contribution from the occupied class is artificially inflated by weighting each occupied example by a hyper-parameter $\alpha$. Note that in the partially observed region $\bm{o}^{u,v}=2$ there is no loss.

\subsection{Inference}
\label{sec:inferenc}
Given a trained model $p_{\varparams_*}(\zz | \xx) = \mathcal{N}(\zz|\bm{\mu}_{\varparams_*}(\xx), \bm{\gamma}_{\varparams_*}(\xx))$ we now wish to determine the probability that each cell is occupied given input $\xx$ by marginalising out the uncertainty associated with the latent variable $\zz$:
\begin{gather}
    p(\yy^{u, v}| \xx) \defeq \int p(\yy^{u, v} | \zz^{u, v}) p_{\varparams_*}(\zz^{u, v} | \xx)d \zz^{u, v}
\end{gather}

However, for likelihood $p(\yy^{u, v} | \zz^{u, v}) = \sigmoid(\zz^{u,v})$ no exact closed form solution exists to this integral. Instead of resorting to Monte Carlo sampling we approximate the sigmoid function with a probit function and use the result that a Gaussian distribution convolved with a probit function is another probit function \cite{nasrabadi2007pattern}. Following this analysis, it can be shown that,
\begin{equation}
     p(\yy^{u, v} = 1| \xx) \approx \sigmoid \bigg(\frac{\bm{\mu}_{\varparams_*}^{u, v}}{\bm{s}_{\varparams_*}^{u, v}} \bigg)
\end{equation}
\noindent where $\bm{s}_{\varparams_*}^{u, v} = (1 + (\bm{\gamma}_{\varparams_*}^{u, v} \sqrt{\pi / 8} )^2)^{1/2}$, $\bm{\mu}_{\varparams_*}^{u, v} = \bm{\mu}_{\varparams_*}^{u, v}(\xx)$ and $\bm{\gamma}_{\varparams_*}^{u, v} = \bm{\gamma}_{\varparams_*}^{u, v}(\xx_*)$. This allows us to efficiently calculate $\bm{p}_{\yy | \xx}$ as,
\begin{gather}
    [\bm{\mu}_{\varparams_*}, \bm{\gamma}_{\varparams_*}] = f_{\varparams_*}(\xx) \\
    \bm{s}_{\varparams_*} = (1 + (\bm{\gamma}_{\varparams_*} \sqrt{\pi / 8} )^2)^{1/2} \\
    \bm{p}_{\yy | \xx} \defeq \sigmoid \bigg(\frac{\bm{\mu}_{\varparams_*}}{\bm{s}_{\varparams_*}} \bigg) \label{eq:analytic_approx}
\end{gather}

\begin{figure}
    \centering
    \includegraphics[width =1\linewidth]{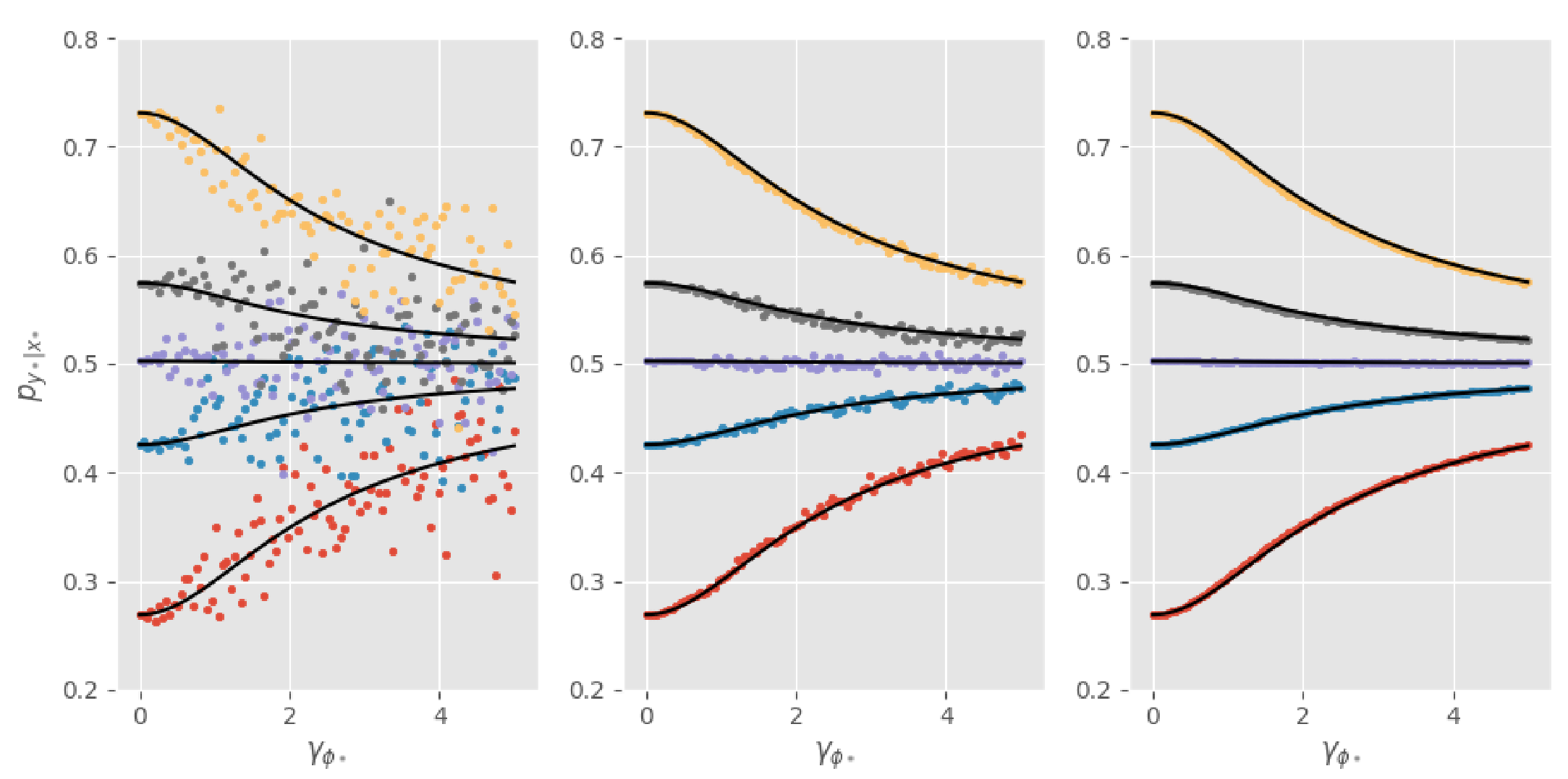}
    \vspace{-0.7cm}
    \caption{Predicted occupancy probabilities $\bm{p}_{\yy | \xx}$ as a function of predicted standard deviation $\bm{\gamma}_{\varparams_*}$ using the analytic approximation given by (\ref{eq:analytic_approx}) (black) vs Monte Carlo approximation with $L=10^2$ (left), $L=10^4$ (middle) and $L=10^6$ (right) samples. Each colour corresponds to a different mean $\bm{\mu}_{\varparams_*}$ with $[\mathsf{yellow}, \mathsf{grey},\mathsf{purple}, \mathsf{blue}, \mathsf{red}] $  corresponding to means $[-1, -0.3, 0.01, 0.3, 1]$ respectively. It is seen that the MC estimate has high variance taking of the order $10^6$ samples to converge to the analytic approximation. On the other hand the analytic approximation closely resembles the converged Monte Carlo estimate.}
    \label{fig:analytic_vs_mc}
    \vspace{-0.4cm}
\end{figure}

Figure \ref{fig:analytic_vs_mc} shows $\bm{p}_{\yy | \xx}$ approximated using (\ref{eq:analytic_approx}) and Monte Carlo sampling for varying $\bm{\mu}_{\varparams_*}$ and $\bm{\gamma}_{\varparams_*}$. The Monte Carlo estimate takes of the order $10^4$ samples to converge, whilst the analytic approximation provides a close approximation to the converged Monte Carlo estimate.

In equation (\ref{eq:analytic_approx}) the predicted logit $\bm{\mu}_{\varparams_*}$ can be thought of as giving the score associated with labelling an example as occupied; intuitively the higher the score the higher the probability that each cell is occupied. In contrast, the predicted deviation  $\bm{\gamma}_{\varparams_*}$ increases the entropy in the predicted occupancy distribution independent of the cells predicted score and captures uncertainties that cannot be easily explained by the predicted score alone.

\section{RESULTS}
\begin{figure}
    \vspace{0.2cm}
    \centering
    \includegraphics[width=\linewidth]{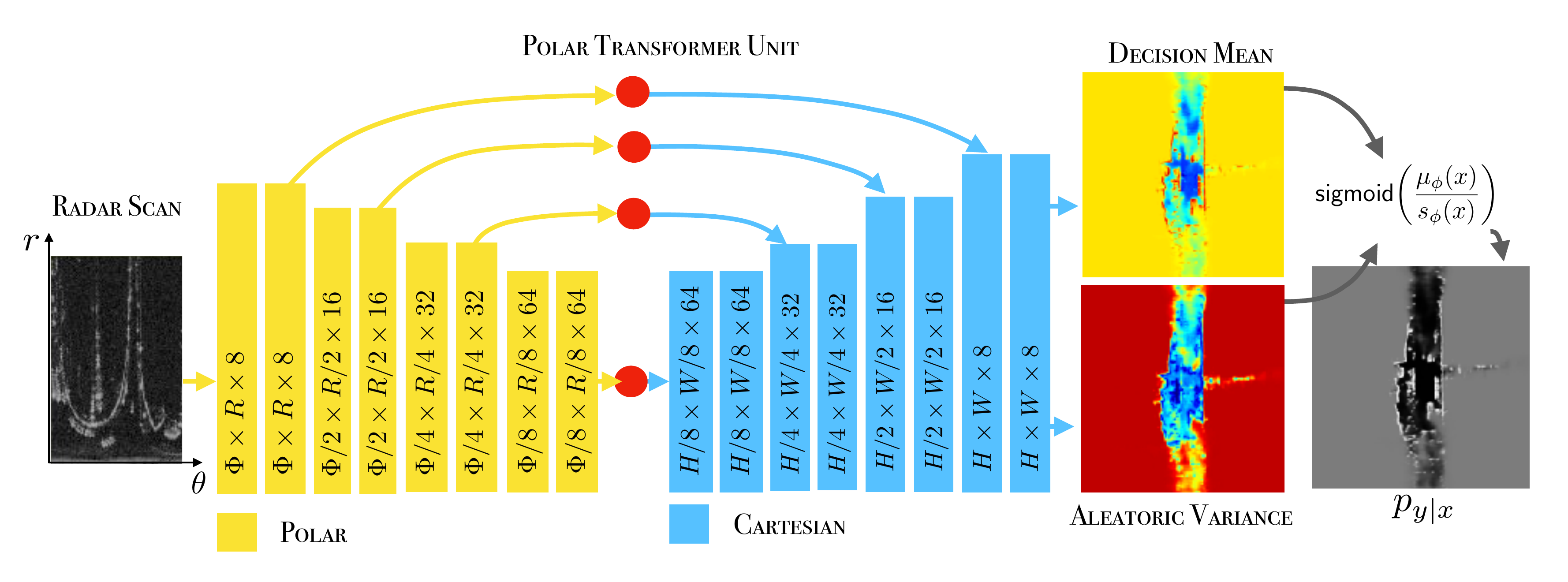}
    \vspace{-0.6cm}
    \caption{Our network architecture takes in a polar radar scan $\xx \in \mathbb{R}^{\Theta \times R}$ and maps it to Cartesian grids of mean utility $\bm{\mu}_\varparams$ and aleatoric noise scale $\bm{s}_\varparams = (1 + (\bm{\gamma}_{\varparams} \sqrt{\pi / 8} )^2)^{1/2}$. Our network is composed of a polar (yellow) encoder and a Cartesian (blue) decoder. At each polar to Cartesian interface there is a polar transformer unit (red circle). Each blue rectangle corresponds to $2$ convolutions followed by a max pool.}
    \label{fig:polnet}
    \vspace{-0.6cm}
\end{figure}

In this Section we show that our model, despite challenging noise artefacts, is able to successfully segment the world into occupied and free space achieving higher mean Intersection over Union (IoU) scores than cell averaging CFAR filtering approaches. In addition to this we are also able to explicitly identify regions of space that are likely to be occluded through the uncertainties predicted by our network. We provide several qualitative examples of our model operating in challenging real world environments and study the effects of our prior on our network output through an ablation study.

\subsection{Experimental Set-Up}
A Navtech CTS350x FMCW radar (without Doppler) and two Velodyne HDL32 lidars were mounted to a survey vehicle and used to generate over $78000$ $(90\%)$ training examples and $8000$ $(10\%)$ test examples from urban scenes. The output from the two lidars was combined from $0.7\text{m}$ below the roof of the vehicle to $1\text{m}$ above and projected onto a $600 \times 600$ grid, with a spatial resolution of $0.3\text{m}$, generating a $180\text{m} \times 180\text{m}$ world occupancy map, following the procedure described in Section \ref{sec:setting}. To account for differences in the frequency of our radar ($4$Hz) and lidar ($10$Hz) the occupancy map was ego-motion compensated such that the Cartesian map corresponds to the time stamps of each radar azimuth.

Figure \ref{fig:polnet} shows our network architecture in which a polar encoder takes the raw radar output and generates a polar feature tensor through repeated applications of $4 \times 4$ convolutions and max pooling before a Cartesian decoder maps this feature tensor to a grid of mean logits $\bm{\mu}_{\varparams}(\xx) \in \mathbb{R}^{H \times W}$ and standard deviations $\bm{\gamma}_{\varparams}(\xx)\in (0, \infty)^{H \times W}$ which are converted to a grid of probabilities through (\ref{eq:analytic_approx}). Information is allowed to flow from the encoder to the decoder through skip connections, where polar features $\bm{u}$ are converted to Cartesian features $\bm{v}$ through bi-linear interpolation, with a fixed polar to Cartesian grid \cite{esteves2017polar}. In all experiments we trained our model using the ADAM optimiser \cite{kingma2014adam}, with a learning rate of $0.001$, batch size $16$ for $100$ epochs and randomly rotated each input output pair about the origin, minimising the loss proposed in (\ref{eq:final_loss}) with $L=25$ samples. Experimentally it was found that setting $\alpha = 0.5$ gave the best results in terms of IoU performance against the lidar labels. Unless otherwise stated, the model evidence importance was set to $\omega=1$.

\begin{figure*}
    \centering
    \subfloat[The detection performance of our approach vs classical filtering methods with black representing predicted free and white representing predicted occupied by each approach. In comparison to CFAR our approach results in crisp and clean detections in observed and unobserved space. The red rectangles highlight cars that are clearly detected by our approach which are largely missed by CFAR. In addition, our model is able to successfully reason about what in the scene is likely to be unknown due to occlusion.]{%
        \centering%
        \includegraphics[width=\linewidth]{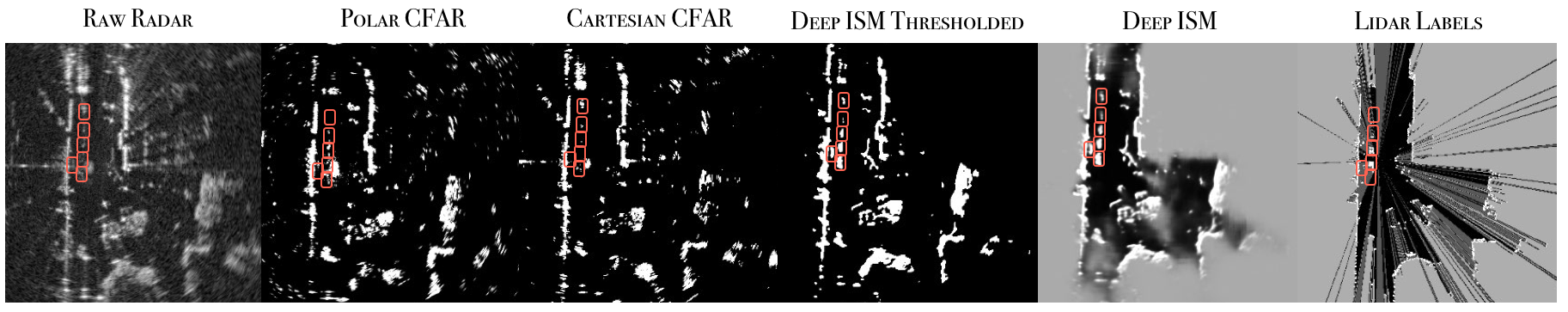}%
        \label{fig:detection_performance}%
    }
    \vspace{-0.02cm}
    \subfloat[The predicted probability of occupancy for different values of likelihood importance $\omega$. As $\omega$ is increased our model becomes increasingly less conservative, reasoning in the unobserved region of space based on labels in the observed region.]{%
        \centering
        \includegraphics[width=\linewidth]{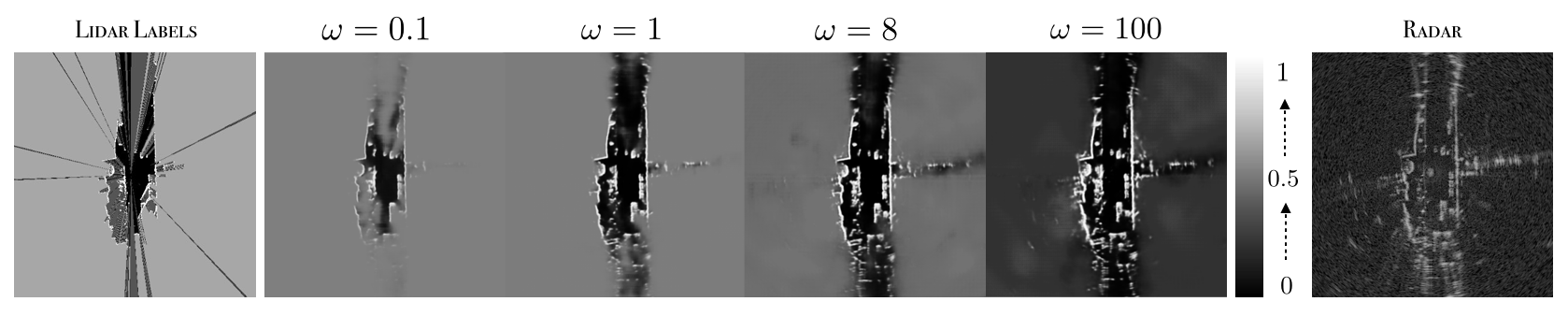}
        \vspace{-0.3cm}
        \label{fig:model_evidence}
        \vspace{0.2cm}%
    }
    \vspace{0.2cm}
    \subfloat[Our model successfully identifies occupied free and occluded space in challenging real world environments.]{
    \centering
        \includegraphics[width=\linewidth]{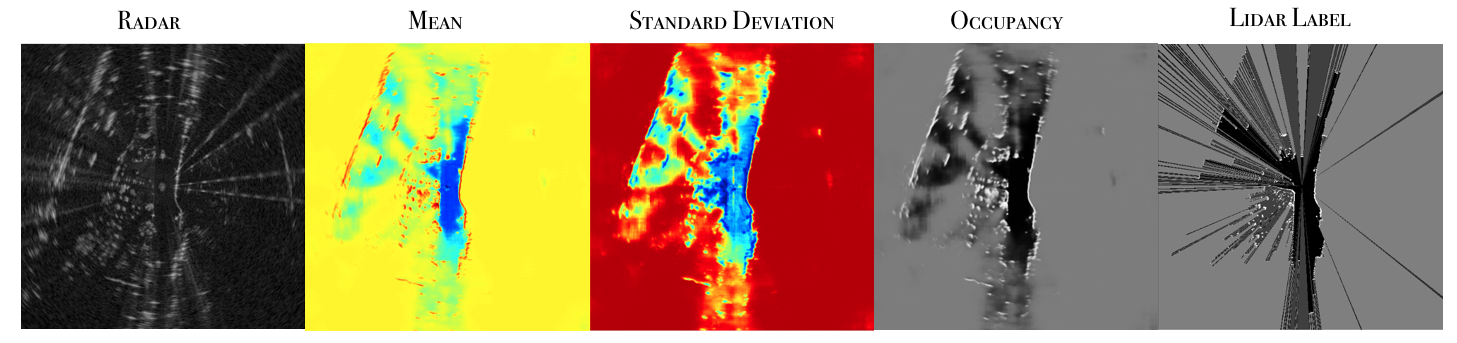}%
        \label{fig:qualitative_results}%
    }
    \vspace{0.2cm}
    \subfloat[A scene segmented as predicted occupied (white), unoccupied (black) and unknown (grey) for decreasing confidence thresholds (left to right) on the predicted standard deviation $\bm{\gamma}_\varparams$.  From most certain to most least certain, we move from high power returns labelled as occupied, to a region nearby and up to the first return, to space that lies in partial and full occlusion.]{
        \centering%
        \includegraphics[width =\linewidth]{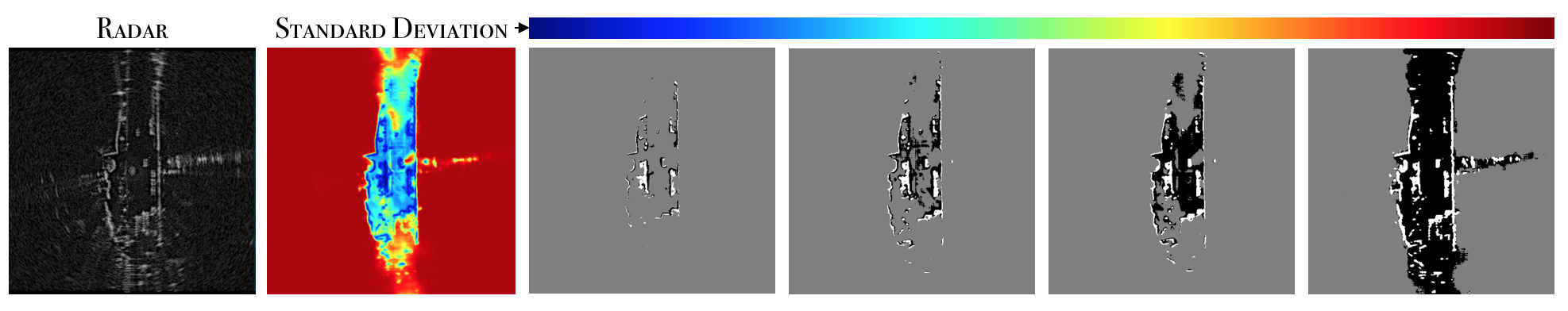}%
        \label{fig:uncertainty_prediction}%
    }
    \caption{}
\end{figure*}

\subsection{Detection Performance of Deep ISM vs Classical Filtering Methods}
We compare the detection performance of our approach against cell averaging CFAR \cite{skolnik2008radar} applied in 1D (along range) for polar scans and in 2D for Cartesian scans by determining the quantity of occupied and unoccupied space successfully segmented in comparison to the ground truth labels generated from lidar in observed space. Due to class frequency imbalance, we use the mean Intersection Over Union (IoU) metric \cite{everingham2010pascal}. The optimum number of guard cells, grid cells and probability of false alarm, for each CFAR method, was determined through a grid search maximising the mean IoU of each approach on training data. For our method, each cell was judged as occupied or free based on a $0.5$ probability threshold on $\bm{p}_{\yy | \xx}$. A $2\text{m}$ square in the centre of the occupancy map, corresponding to the location of the survey vehicle, was marked as unobserved.

The results form the test data set for each approach are shown in table \ref{tab:detection_performance} and show that our approach outperforms all the tested CFAR methods, increasing the performance in occupied space by 0.11, whilst achieving almost the same performance in free space leading to a mean IoU of $0.63$. Our model is successfully able to reason about occupied space in light of challenging noise artefacts. In contrast, the challenge in free space is not in identification, with free space typically being characterised by low power returns, but in distinguishing between observed and occluded regions, a challenge which is missed entirely by the IoU metric. Figure \ref{fig:detection_performance} shows how our model is able to successfully determine space that is likely to be unknown because of occlusion and is able to clearly distinguish features, such as cars that are largely missed in CFAR. An occupancy grid of size $600 \times 600$ can be generated at around $14$Hz on a NVIDIA Titan Xp GPU. Which is significantly faster than real time for radar with a frequency of $4$Hz.

\begin{table}
\vspace{0.3cm}
\begin{center}
\caption{Comparing our approach to classical detection methods using Intersection over Union}\label{tab:detection_performance}
\begin{tabularx}{\columnwidth}{@{\extracolsep{\fill}}@{ } l c c c}
\toprule
\vspace{0.5pt}
 & \multicolumn{3}{c}{Intersection over Union}  
\\
 \cmidrule(lr){2-4} 
  Method & Occupied & Free & Mean \\
 \midrule
 CFAR (1D polar) &  0.24     & \textbf{0.92} & 0.5
 \\
 CFAR (2D Cartesian) & 0.20     & 0.90 & 0.55 \\
 Static thresholding & 0.19     & 0.77 & 0.48 \\
 Deep ISM (our approach) & \textbf{0.35}     & 0.91 & \textbf{0.63}  \\
\bottomrule
\end{tabularx}\vspace{-2em}
\end{center}
\vspace{-0.2cm}
\end{table}

\subsection{Uncertainty Prediction}
\label{sec:uncertainty_prediction}

As described in Section \ref{sec:inferenc}, by incorporating aleatoric uncertainty into our formulation, the latent uncertainty associated with each grid cell is allowed to vary by predicting the standard deviation of each cell $\bm{\gamma}_{\varparams}(\xx)$ alongside the predicted logit $\bm{\mu}_{\varparams}(\xx)$. In this section we investigate the uncertainties that are captured by this mechanism.

To do this we gradually increase a threshold on the maximum allowable standard deviation of each cell $\bm{\gamma}_{\varparams}(\xx)$ labelling any cell that falls below this threshold as either occupied (white) or free (black), whilst every cell above the threshold is labelled as unknown (grey). The result of this process is illustrated in Figure \ref{fig:uncertainty_prediction}.

The standard deviation predicted by our network largely captures uncertainty caused by occlusion, which, independent of the true underlying state of occupancy, results in space that is inherently unknown. From least likely to most likely to be occluded, we move from high power returns labelled as occupied, to a region nearby and up to the first return, to space that lies in partial and full occlusion. This ray tracing mechanism is largely captured by the standard deviation $\bm{\gamma}_{\varparams}(\xx)$ predicted by our network.

\subsection{Qualitative Results}
Finally, we provide several qualitative examples of our model operating in challenging real world environments and investigate how the strength of our prior term in (\ref{eq:final_loss}) effects the occupancy distribution predicted by our model.

Figure \ref{fig:qualitative_results} gives qualitative examples taken from the test set. Our network is able to successfully reason about the complex relationship between observed and unobserved space in light of challenging noise artefacts. In Figure \ref{fig:model_evidence} we vary the relative importance between the likelihood and KL divergence term by varying the hyper-parameter $\omega$ in (\ref{eq:final_loss}). Increasing $\omega$ increases the relative importance of the likelihood term and leads to an ISM which is able to more freely reason about regions of space for which no labels exist during training, using the labels available in the observed scene. In the limit, of high $\omega$ the model is no longer able to successfully identify regions of space that are likely to be occluded, predicting all low power returns as free with a high probability.

                                  
\section{CONCLUSION}
By using a deep network we are able to learn an inherently probabilistic ISM from raw radar data that is able to identify regions of space that are likely to be occupied in light of complex interactions between noise artefacts and occlusion. By accounting for scene context, our model is able to outperform CFAR filtering approaches. Additionally, by modelling heteroscedastic uncertainty we are able to capture the variation of uncertainty throughout the scene, which can be used to identify regions of space that are likely to be occluded. Our network is self-supervised using only partial labels generated from a lidar, allowing a robot to learn about the occupancy of the world by simply traversing an environment. 

At present our approach operates under a static world assumption. In future work we hope to incorporate scene dynamics into our formulation allowing a robot to identify cells that are likely to be dynamic in addition to occupied or free.




\section*{ACKNOWLEDGMENT}
The authors would like to thank Oliver Bartlett and Jonathan Attias for proof reading a draft of the paper, and Dan Barnes for many insightful conversations, and the reviewers for helpful feedback. 

This work was supported by training grant Programme Grant EP/M019918/1. We acknowledge use of Hartree Centre resources in this work. The STFC Hartree Centre is a research collaboratory in association with IBM providing High Performance Computing platforms funded by the UK’s investment in e-Infrastructure. The Centre aims to develop and demonstrate next generation software, optimised to take advantage of the move towards exa-scale computing.


\newpage
\bibliography{bib}
\bibliographystyle{ieeetr}




\end{document}